\documentclass[10pt,twocolumn,letterpaper]{article}

\usepackage{cvpr}              

\usepackage{amsmath} 

\definecolor{cvprblue}{rgb}{0.21,0.49,0.74}
\usepackage[pagebackref,breaklinks,colorlinks,allcolors=cvprblue]{hyperref}
\usepackage{graphicx} 
\usepackage{amsmath}   
\usepackage{amssymb}   
\usepackage{amsfonts}  
\usepackage{bm}        

\usepackage{hyperref}  
\usepackage{colortbl}
\usepackage{multirow}
\usepackage{bbding}
\usepackage{amssymb}
\usepackage{amsmath}
\usepackage{pifont} 
\usepackage{xcolor}

\definecolor{secondcolor}{RGB}{189,215,238}
\definecolor{firstcolor}{RGB}{255,153,153}
\newcommand{\firstcolor}[1]{\cellcolor[rgb]{1,.60,.60}{#1}}
\newcommand{\secondcolor}[1]{\cellcolor[rgb]{.741,.843,.933}{#1}}

\usepackage{algorithm}
\usepackage{algpseudocode}
\def\paperID{} 
\def\confName{CVPR}
\def\confYear{2026}

\title{Unpaired Image Deraining Using Reward-Guided Self-Reinforcement Strategy}

\author{
Yinghao Chen$^{1}$\quad %
Yeying Jin$^{2}$\textsuperscript{$\dagger$,$\ddag$}\quad %
Xiang Chen$^{3}$\quad %
Yanyan Wei$^{4}$\textsuperscript{$\dagger$}\quad %
Ziyang Yan$^{5}$\quad 
Yaowen Fu$^{1}$\\
{\small $^{1}$National University of Defense Technology} %
{\small \quad$^{2}$ National University of Singapore}\\
{\small $^{3}$Nanjing University of Science and Technology} %
{\small \quad$^{4}$ Hefei University of Technology} %
{\small \quad$^{5}$University of Trento}
}
\begin{document}

\twocolumn[{%
\renewcommand\twocolumn[1][]{#1}%
\maketitle
\begin{center}
    \centering
    \captionsetup{type=figure}
    \vspace{-6mm}
    \includegraphics[width=\textwidth]{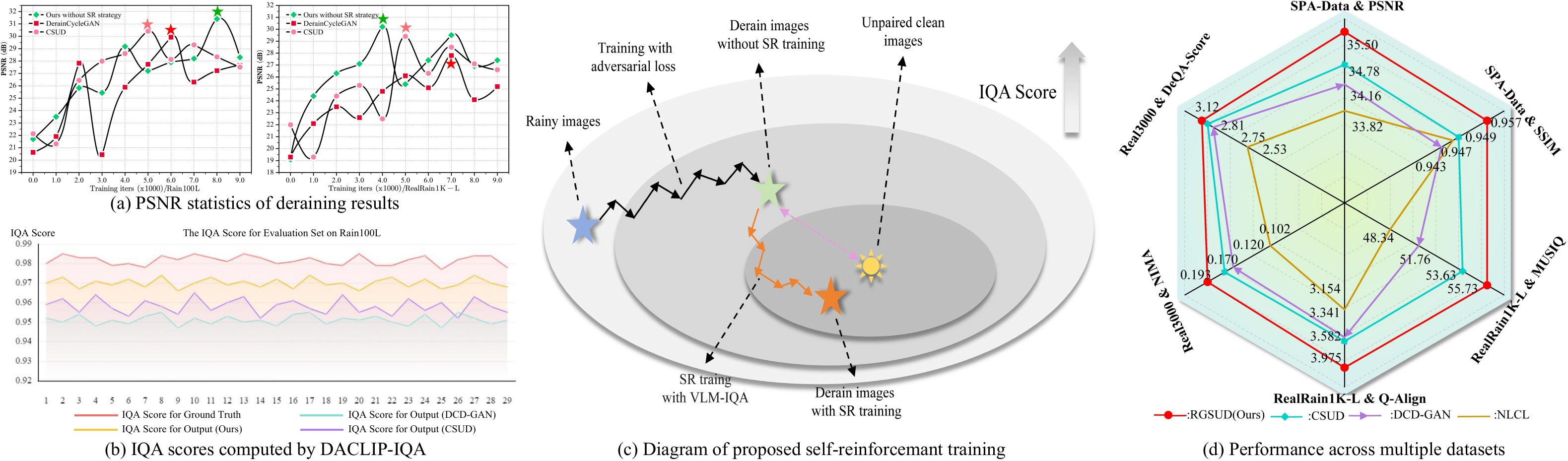}
    \vspace{-18pt}
    \caption{\textbf{Observation, Motivation, Methodology, and Performance of Our Work.} (a) PSNR statistics of deraining results during unsupervised training on Rain100L \cite{yang2017deep} and RealRain1K-L \cite{li2022toward} datasets, showing the observed performance during training. (b) Image Quality Assessment (IQA) scores from the Vision Language Model (VLM)-based DACLIP-IQA \cite{luo2024controlling} on the Rain100L dataset. VLM-based IQA provides perceptual scores as rewards for self-reinforcement (SR) optimization. (c) The schematic diagram of our RGSUD algorithm, where the stars $\bigstar$ represent the evolution of sample latent states during training. We propose an SR strategy to enhance deraining results by improving perceptual quality. (d) The results of RGSUD demonstrate significant improvements across multiple datasets, including paired synthetic and real images and unpaired real images, achieving SOTA results in subjective, and no-reference IQA metrics.
    \label{fig:teaser}} 
    \vspace{-0.5em}
\end{center}%
}]

{
  \renewcommand{\thefootnote}{\fnsymbol{footnote}} 
  \footnotetext[2]{Corresponding Authors.} 
  \footnotetext[3]{Project Lead.} 
}
\begin{abstract}
Unsupervised deraining has attracted attention for its ability to learn the real-world distribution of rain without paired supervision. However, the lack of strong constraints makes it difficult for the network to converge, especially with the complex diversity of rain degradation. A key motivation is that high-quality deraining results occasionally emerge during training, which can be leveraged to guide the optimization process. To overcome these challenges, we introduce \textbf{RGSUD} (Reward-Guided Self-Reinforcement Unsupervised Image Deraining), comprising two key stages: reward recycling and self-reinforcement (SR) training. For the former stage, we propose an Image Quality Assessment (IQA)-based dynamic reward recycling mechanism that selects optimal derained outputs during training and continuously collects high-quality deraining images. In latter stage, we incorporate these rewards into the model’s optimization process, constraining the optimization space and improving alignment between derained outputs and clean images. By leveraging IQA-based self-reinforced loss and dynamically updated rewards, we enhance the quality of synthesized pseudo-paired data and stabilize the optimization. Extensive experiments demonstrate that our method achieves SOTA performance across multiple datasets, including paired synthetic, paired real, and unpaired real images, outperforming existing unsupervised deraining approaches in both subjective and objective IQA metrics. Additionally, we show that the Self-Reinforcement Strategy is adaptable to other unsupervised deraining methods and our deraining framework demonstrates strong generalization across existing supervised deraining networks.
\end{abstract}    
\section{Introduction}
\label{sec:intro}

Rain severely degrades the performance of high-level vision tasks, such as object detection \cite{liu2025causal,zhang2025advancing,10444566,10898024,yan2025renderworld,wang2025unifying} and recognition \cite{li2025saratr,su2025rapid,yan20243dsceneeditor}, highlighting the need for robust image deraining techniques. While supervised deraining methods have made significant strides with the development of neural networks~\cite{kong2025luminanceawarestatisticalquantizationunsupervised,guo2024onerestore,li2025instruct2see}, they are heavily reliant on synthetic-clean image pairs and often struggle to generalize to real-world rainy patterns due to domain discrepancies \cite{ye2022unsupervised}. In contrast, unsupervised deraining methods directly learn distributions of rain degradation from real-world data, offering better adaptability to diverse, realistic rainfall scenarios. However, training effective deraining networks using unpaired clear and rainy images remains challenging due to the absence of explicit constraints for both domains, leading to under-constrained optimization \cite{chen2022unpaired}.

A key observation from our experiments (Figure \ref{fig:teaser} (a)) is that high-quality deraining results often emerge during the training process, suggesting that these intermediate outputs may provide implicit supervision for deraining. This insight motivates the question: \textit{\textbf{Can we recycle high-quality intermediate results during training as rewards to regularize learning and compact the optimization space?}} Inspired by reinforcement learning, where reward mechanisms guide policies toward optimal behavior, we propose using high-quality deraining outputs as \textit{rewards} to guide the optimization process. However, the challenge lies in managing the diversity of deraining results during training and in designing a reliable, generalized reward recycling mechanism. Figure \ref{fig:teaser} (b) illustrates this approach, showing how perceptual quality scores, computed using the VLM-based DACLIP-IQA \cite{luo2024controlling}, can serve as reliable rewards during training. IQA scores effectively distinguish degraded and derained images across varying levels of fidelity, providing a more accurate and consistent measure of perceptual quality than traditional metrics such as PSNR. By using these perceptual scores as rewards, we can more effectively guide the optimization of deraining models, improving their ability to generate high-quality outputs aligned with human visual perception. This leads us to introduce a self-reinforcement (SR) mechanism that recycles perceptual rewards to regularize learning, compact the optimization space, and promote convergence toward high-fidelity deraining results~\cite{hu2025omniview,hu2026geometry,hu2025auto,yan2025synthetic,yan2026comp}.

To overcome this challenge, we propose \textbf{RGSUD}, a reward-guided self-reinforcement unsupervised image deraining framework. Central to our approach is a dynamic reward recycling mechanism based on a VLM-based IQA, which selects optimal deraining results as rewards during model training. As shown in Figure \ref{fig:teaser} (c), we introduce a two-stage unpaired deraining training paradigm. In the \textit{recycling stage}, the network is trained on unpaired data, while high-quality outputs are dynamically recycled as rewards, detached from the backpropagation graph. In the \textit{self-reinforcement stage}, these rewards are incorporated into the gradient flow in two ways: (i) guiding the synthesis of high-quality pseudo-paired rainy images and (ii) constructing a self-reinforcement loss to restrict the optimization space. The first path uses dynamically updated rewards to refine pseudo-paired data and the reconstruction process, progressively improving the generated data. The second path introduces a novel loss term that aligns the distributions of derained images with those of natural images, leading to more consistent and reliable results. Our main contributions are summarized as follows:

\begin{itemize}
\item We introduce a novel reward-guided self-reinforcement unsupervised image deraining framework (\textbf{RGSUD}) that effectively exploits implicit supervision and VLM knowledge to mitigate the challenge of a lack of paired supervision in various rain scenarios.

\item We propose a dynamic \textbf{Reward Recycling Mechanism} that leverages a pretrained VLM to evaluate and select high-quality intermediate deraining results as rewards, marking its first application in the unsupervised deraining field.

\item By combining the Reward Recycling Mechanism and a \textbf{Degradation Estimation Module}, we introduce a \textbf{Self-Reinforcement Strategy}, which helps the network converge to a compact optimization space, thereby enhancing the distribution alignment between deraining outputs and natural images.

\item Extensive experiments demonstrate that our method achieves SOTA performance across multiple datasets, including paired synthetic images, paired real images, and unpaired real images, and outperforms existing unsupervised deraining approaches in both subjective and objective IQA metrics (Figure \ref{fig:teaser} (d)). Besides, we also demonstrate that this Self-Reinforcement Strategy can be adaptable to other unsupervised deraining methods.
\end{itemize}
\section{Related Work}

\noindent {\textbf{\ding{113} Single Image Deraining.}}
There has been a proliferation of deraining baselines designed to support learning from paired data, including CNN-based frameworks (\textit{e.g.}, MSPFN \cite{jiang2020multi}), Transformer-based solutions (\textit{e.g.}, DRSformer \cite{chen2023learning}), and the latest Mamba-based architectures (\textit{e.g.}, RainMamba \cite{wu2024rainmamba}). Nonetheless, the scarcity of large-scale paired training data in real-world scenarios hinders their application{~\cite{jin2025raindrop,li2025ntire}}.

To address the lack of paired datasets in the real world, some remarkable unsupervised{~\cite{jin2021dc,jin2022unsupervised,zheng2024orientation}} deraining approaches have been proposed~\cite{dong2026learningdomainawaretaskprompt}. Wei \textit{et al.} \cite{wei2021deraincyclegan} proposed an unsupervised method by utilizing the CycleGAN \cite{zhu2017unpaired} framework. 
Ye \textit{et al.} \cite{ye2022unsupervised} proposed an unsupervised non-local contrastive learning deraining method to more effectively separate rain from the image. 
Chen \textit{et al.}~\cite{chen2022unpaired} strive to constrain the optimization space from a feature space perspective by leveraging the contrastive learning paradigm. 
Recently, Dong \textit{et al.} \cite{dong2025channel} explored the channel consistency prior for rainy images, demonstrating strong robustness. Despite these advancements, aligning with ground truths remains a significant challenge{~\cite{remondino2023critical,yan2023nerfbk}}.

\noindent {\textbf{\ding{113} VLM-based IQA Models.}}
VLM-based IQA methods leverage VLMs' foundational knowledge to achieve better performance \cite{luo2024controlling,wu2023q1}. E.g., Q-align converts score labels into five discrete levels to train the model, then uses Softmax pooling to obtain the probability of each level, and gets the final score through weighted summation, resulting in more accurate score regression. Dog-IQA \cite{liu2024dog} incorporates specific standards and local semantic objects. DeQA-Score \cite{you2025teaching} discretizes the score into level tokens and calculates their respective probabilities to form a soft label. DACLIP-IQA \cite{luo2024controlling} generates quality scores by using softmax to predict the quality level between text features and image encoding features, consistently outperforming other methods in score regression{~\cite{shao2025eventvad,shao2025accidentblip,shao2025icm,jiang2026medical}}.

\noindent {\textbf{\ding{113} VLM-IQA for Image Restoration.}}
Recent research proves that incorporating external priors from pretrained VLMs can significantly improve the performance of image restoration tasks \cite{xu2025enhancingdiffusionbasedrestorationmodels}{~\cite{liao2025gm,yan2025learning}}.
Cheng \textit{et al.} \cite{cheng2024transfer} implemented CLIP within a trainable image decoder to facilitate adaptable image denoising. Leveraging CLIP’s semantic-versatile nature, Jiang \textit{et al.} \cite{jiang2024autodir} created the comprehensive AutoDIR for automated image restoration. Lin \textit{et al.} \cite{lin2025jarvisir,chen2025genhaze} used a VLM-powered agent to manage and coordinate multiple expert restoration models to address coupled weather degradations. However, their applicability in unpaired image deraining remains underexplored~\cite{lin2024dual}.
\newcommand{\vect}[1]{\bm{#1}} 
\newcommand{\matr}[1]{\bm{#1}} 
\newcommand{\norm}[1]{\left\lVert#1\right\rVert} 
\newcommand{\calD}{\mathcal{D}} 
\newcommand{\calF}{\mathcal{F}} 
\newcommand{\calX}{\mathcal{X}} 
\newcommand{\calO}{\mathcal{O}} 
\newcommand{\calQ}{\mathcal{Q}} 

\begin{figure*}[!t]
	\centering
	\includegraphics[width=1.0\textwidth]{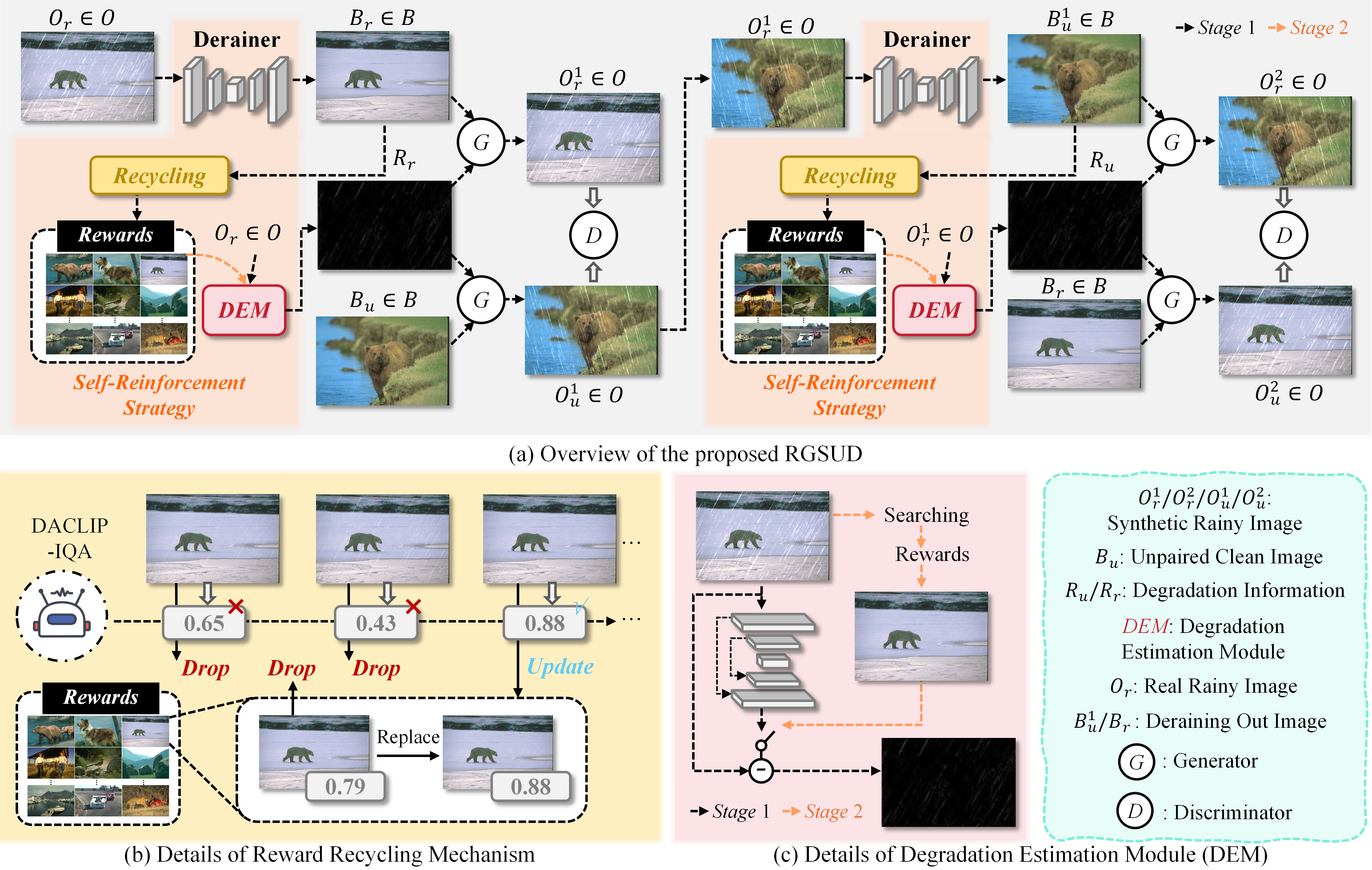}
	\vspace{-6mm}
    \caption{The framework of the proposed RGSUD. (a) Overview of the network architecture, illustrating the Reward Recycling and Self-Reinforcement Strategy. (b) Details of the Reward Recycling Mechanism, which evaluates and updates high-quality derained outputs using the VLM-based IQA. (c) Detailed architecture of the Degradation Estimation Module (DEM), which decides whether to enable rewards for synthesizing pseudo-paired data during training.}
    	\label{fig2}
	\vspace{-1.5em}
\end{figure*}

\section{Methodology}
\label{sec:methodology}

\subsection{Overall Framework}
As shown in Figure \ref{fig2} (a), our RGSUD mainly comprises four components: Derainer, Degradation Estimation Module (DEM), generator ${G}$ and Discriminator ${D}$. Derainer employs NAFNet~\cite{chen2022simple} for removing rain. DEM consists of a simple  U-Net~\cite{ronneberger2015u} to obtain rain information and a switch that controls whether the reward participates in exacting rain information. 
We employ a ResNet-based generator ${G}$ with six residual blocks, and PatchGAN~\cite{zhu2017unpaired} as the discriminator ${D}$.
What's more, the `` Recycling'' in Figure \ref{fig2} (a) represents the proposed VLM-based dynamic reward recycling mechanism, which is briefly illustrated in Figure \ref{fig2} (b). 

In generator ${G}$, we effectively leverage unpaired clean and rainy images to synthesize pseudo-paired images. As shown in Figure \ref{fig2} (d), we design a GAN-based baseline that transitions from ``$B_u\in B \rightarrow O_u^1\in O \rightarrow B^{1}_u\in B $'', where $B, O$ denote clean images and rainy images. The process ``${G}(B_u,\mathcal{F}(O_r))\rightarrow O_u^{1}$ '' transforms clean image $B_u$ to rainy image $O_u^1$ according to rain information of $O_r$, where ``$\mathcal{F}(\cdot)$'' is the DEM. 
Specifically, to mitigate the challenge of content mismatch in unpaired data, which exacerbates the difficulty of the generator, a rainy image $O_r$ is fed into the U-Net in the DEM to extract a clean image feature, and the rain information is obtained via residual computation. 

To train the RGSUD effectively, we divide the training process into two stages, as illustrated in Figure \ref{fig2} (a). 
In stage one, our data flow is shown by the black arrows, and we utilize the ``Recycling'' to obtain the high-quality deraining images as rewards during model training. And the switch in DEM is moved to the black endpoint. At this stage, the rewards are detached from the backpropagation graph.  
In stage two, our data flow is shown by the black arrows and orange arrows. Building upon the network weights from stage one, we continue training by introducing the self-reinforcement constraint and moving the switch in DEM to the orange endpoint. At this stage, the rewards are also dynamically updated, continuing until the framework's deraining performance ceases to improve. Please refer to the supplementary for more detailed algorithms and structures. Next, we describe the Degradation Estimation Module and Reward Recycling Mechanism~\cite{fang2024real,fangphoton,fang2025integrating,he2023reti,he2025diffusion}.


\subsection{Reward Recycling Mechanism}
In this section, we will introduce the specific steps of the reward recycling mechanism.
We aim to use it to obtain rewards that guide rain removal. Specifically, we use ``Recycling'', which introduces a robust VLM-based dynamic reward recycling mechanism to obtain rewards during the model training. Below, we introduce the dynamic reward recycling mechanism~\cite{yao2025polar,yao2024neural,yao2023generalized,zhang2023ingredient}.

\noindent {\textbf{\ding{113} Dynamic Reward Recycling.}}
Notably, expansive VLMs, pretrained on large datasets , excel in image quality evaluation (IQA) \cite{jiang2024autodir}. Therefore, we leverage the inherent zero-shot abilities of these VLMs \cite{luo2024controlling} to assess reconstructed training samples. The rewards updating algorithm based on DACLIP-IQA is illustrated in Algorithm \ref{alg1} and Figure \ref{fig2} (b). Specifically, for the rainy image dataset $D_r = \{x_i\}_{i=1}^{N}$, we get the deraining image $x_i^{rec}$ from ``$x_i^{rec}=D_{er}(x_i)$'', where $D_{er}(\cdot)$ is the Derainer in Figure \ref{fig2}. Then, we determine the update and assignment of rewards $\mathcal{R\{\cdot\}} $ based on the scores from DACLIP-IQA $\Psi(\cdot)$.


\vspace{-0.5em}
\begin{algorithm}
\caption{The Process of Dynamic Reward Update}
\label{alg1}
\begin{algorithmic}[1]
\State {\bf Require:} IQA Method $\Psi(\cdot)$, rainy image dataset $D_r = \{x_i\}_{i=1}^{N}$, the Derainer: $D_{er}(\cdot)$, IQA score: $z$;
\State Initialize rewards $\mathcal{R}\{\cdot\} = \emptyset$
\For{each $x_i$}
    \State Rain removal process: {$x_i^{rec}=D_{er}(x_i)$};
    \State Compute IQA scores of $x_{i}^{rec}$, $x_{i}^{r}\in{\mathcal{R}\{\cdot\}}$;
    \State $z_{rec}=\Psi(x_i^{rec})$, $z_{r} \Psi(x_i^{r})$;
    \If {$z_{rec} > z_{r}$}
        \State Replace $x_i^{r} $ in $\mathcal{R}\{\cdot\}$ by $x_i^{rec}$;
    \Else
        \State{Keep $x_i^r$ in $\mathcal{R}\{\cdot\}$}
    \EndIf
\EndFor
\end{algorithmic}
\end{algorithm}
\vspace{-1.5em}

\subsection{Self-Reinforcement Strategy}
In this subsection, we will introduce how DEM leverages rewards to synthesize higher-quality pseudo-paired data. Our key insight is that leveraging more accurate rain information enables the synthesis of higher-quality pseudo-paired data, thereby helping improve the performance of deraining networks through the loss function \ref{eq6}, which forms a gain loop~\cite{lin2023unsupervised,lin2025re,lin2023multi}.

\noindent {\textbf{\ding{113} Degradation Estimation Module.}} 
As shown in Figure \ref{fig2} (c), we adopt reward as clean feature representations within the DEM, bypassing U-Net-based feature extraction. This approach consistently yields more accurate rain information during residual computation, as the rewards reflect the optimal deraining outputs produced by the NAFNet derainer, which demonstrates superior image restoration capabilities compared to the U-Net. Thus, progressively optimized rewards enable more accurate rain information, which in turn enables the synthesis of higher-quality pseudo-paired data and enhances deraining performance, thereby establishing a stable and reliable cycle.

\begin{table*}[!htb]
    \renewcommand{\arraystretch}{1}
    \centering
    \caption{Quantitative deraining performance comparisons of different methods (\textit{i.e.,} \textcolor{gray}{{\textit{\textbf{Supervised Methods}}}} and \textcolor{gray}{{\textit{\textbf{Unsupervised Methods}}}}) on various rain datasets (\textit{i.e.,} \textcolor{gray}{{\textit{\textbf{Synthetic Paired Datasets}}}} and \textcolor{gray}{{\textit{\textbf{Real-world Paired Datasets}}}}) using \textcolor{gray}{\textit{\textbf{Object Reference IQA Metrics}}} (\textit{i.e.,} PSNR and SSIM). Notably, the best value in each column is highlighted in \colorbox{firstcolor}{red}, and the second best in \colorbox{secondcolor}{blue}.}
    \vspace{-3mm}
    \resizebox{1.0\linewidth}{!}{
    \begin{tabular}{lcccccccccccccc}
        \toprule
\multirow{2}{*}{\textbf{Datasets}} & \multicolumn{8}{c}{\textcolor{gray}{{\textit{\textbf{Synthetic Paired Datasets}}}}} & \multicolumn{6}{c}{\textcolor{gray}{{\textit{\textbf{Real-world Paired Datasets}}}}} \\ 
\cmidrule(lr){2-9}
\cmidrule(lr){10-15}
~ & \multicolumn{2}{c}{Rain100L \cite{yang2017deep}}      & \multicolumn{2}{c}{Rain200L \cite{yang2017deep}}      & \multicolumn{2}{c}{DID-Data \cite{zhang2018density}}      & \multicolumn{2}{c}{DDN-Data \cite{liu2020deep}} & \multicolumn{2}{c}{SPA-Data \cite{wang2019spatial}}      & \multicolumn{2}{c}{RealRain1K-L \cite{li2022toward}}      & \multicolumn{2}{c}{Night-Rain \cite{11075607}} \\ \midrule
Metrics                   & PSNR$\uparrow$ & \multicolumn{1}{c}{SSIM$\uparrow$}  & PSNR$\uparrow$ & \multicolumn{1}{c}{SSIM$\uparrow$} & PSNR$\uparrow$ & \multicolumn{1}{c}{SSIM$\uparrow$} & PSNR$\uparrow$         & \multicolumn{1}{c}{SSIM$\uparrow$}         & PSNR$\uparrow$ & \multicolumn{1}{c}{SSIM$\uparrow$} & PSNR$\uparrow$ & \multicolumn{1}{c}{SSIM$\uparrow$} & PSNR$\uparrow$           & SSIM$\uparrow$          \\ \midrule
\multicolumn{15}{l}{\textcolor{gray}{{\textit{\textbf{Supervised Methods}}}}}\\ 
DDN \cite{fu2017removing} {\scriptsize\textcolor{gray}{[CVPR2017]}} & 32.38          & \multicolumn{1}{c}{0.926}          & 34.68 & \multicolumn{1}{c}{0.844} & 30.97 & \multicolumn{1}{c}{0.912} & 30.00         & \multicolumn{1}{c}{0.904}         & 36.16 & \multicolumn{1}{c}{0.946} & {31.18}   & \multicolumn{1}{c}{0.917}     & {32.43}             & {0.926}                \\
SPANet \cite{wang2019spatial} {\scriptsize\textcolor{gray}{[CVPR2019]}} & 31.95          & \multicolumn{1}{c}{0.919}          & 35.79 & \multicolumn{1}{c}{0.965} & 33.04 & \multicolumn{1}{c}{0.949} & 29.85         & \multicolumn{1}{c}{0.912}         & 40.24 & \multicolumn{1}{c}{0.981} & {30.43}   & \multicolumn{1}{c}{0.947}     & {31.10}             & {0.925}                \\
MPRNet \cite{9577298} {\scriptsize\textcolor{gray}{[CVPR2021]}} & 34.95          & \multicolumn{1}{c}{0.959}          & 39.47 & \multicolumn{1}{c}{0.983} & 33.99 & \multicolumn{1}{c}{0.959} & 33.10         & \multicolumn{1}{c}{0.935}         & 43.64 & \multicolumn{1}{c}{0.984} & {36.29}   & \multicolumn{1}{c}{0.972}     & {36.63}             & {0.966}                \\
Restormer \cite{zamir2022restormer} {\scriptsize\textcolor{gray}{[CVPR2022]}} & 37.57          & \multicolumn{1}{c}{0.974}          & 40.99 & \multicolumn{1}{c}{0.989} & 35.29 & \multicolumn{1}{c}{0.964} & 34.20         & \multicolumn{1}{c}{0.957}         & 47.98 & \multicolumn{1}{c}{0.992} & {40.90}   & \multicolumn{1}{c}{0.985}     & {36.92}             & {0.969} \\
DRSformer \cite{chen2023learning} {\scriptsize\textcolor{gray}{[CVPR2023]}} & {41.23}        & \multicolumn{1}{c}{0.983}          & 41.23 & \multicolumn{1}{c}{0.989} & 35.35 & \multicolumn{1}{c}{0.965} & 34.35         & \multicolumn{1}{c}{0.959}         & 48.54 & \multicolumn{1}{c}{0.992} & {38.84}   & \multicolumn{1}{c}{0.982}     & {37.86}             & {0.971}                \\
NeRD-Rain-S \cite{chen2024bidirectional} {\scriptsize\textcolor{gray}{[CVPR2024]}} & 42.00          & \multicolumn{1}{c}{0.990}          & 41.30 & \multicolumn{1}{c}{0.990} & 35.36 & \multicolumn{1}{c}{0.965} & 34.25         & \multicolumn{1}{c}{0.958}         & 48.90 & \multicolumn{1}{c}{0.994} & {38.64}   & \multicolumn{1}{c}{0.979}     & {38.42}             & {0.973}                \\ \midrule
\multicolumn{15}{l}{\textcolor{gray}{{\textit{\textbf{Unsupervised Methods}}}}}\\ 
CycleGAN \cite{zhu2017unpaired} {\scriptsize\textcolor{gray}{[ICCV2017]}}                         & 29.61          & \multicolumn{1}{c}{0.854}          & 28.84 & \multicolumn{1}{c}{0.874} & 27.76 & \multicolumn{1}{c}{0.838}                   & 27.98         & \multicolumn{1}{c}{0.842}         & 33.54 & \multicolumn{1}{c}{0.913}                   & 20.19 & \multicolumn{1}{c}{0.820} & 21.50           & 0.783          \\
DerainCycleGAN \cite{wei2021deraincyclegan} {\scriptsize\textcolor{gray}{[TIP2021]}} & 32.31 & 0.946 & 31.79 & 0.913 & 28.43 & \secondcolor{0.864} & 28.53 & 0.870 & 34.12 & \secondcolor{0.950} & 28.16 & 0.901 & 27.84 & 0.851 \\
NLCL \cite{ye2022unsupervised} {\scriptsize\textcolor{gray}{[CVPR2022]}} & 27.86          & \multicolumn{1}{c}{0.852}          & 26.91 & \multicolumn{1}{c}{0.813} & 25.89 & \multicolumn{1}{c}{0.813}                   & 26.26         & \multicolumn{1}{c}{0.821}         & 33.82 & \multicolumn{1}{c}{0.947}                   & 23.06 & \multicolumn{1}{c}{0.832} & 25.58           & 0.821         \\
DCD-GAN \cite{chen2022unpaired} {\scriptsize\textcolor{gray}{[CVPR2022]}} & 31.82          & \multicolumn{1}{c}{0.941}          & 31.37 & \multicolumn{1}{c}{0.934} & 28.64 & \multicolumn{1}{c}{0.862}                   & 28.66         & \multicolumn{1}{c}{0.878}         & 34.16 & \multicolumn{1}{c}{0.943}                   & 30.49 & \multicolumn{1}{c}{0.939} & 28.68           & 0.867          \\
CSUD \cite{dong2025channel} {\scriptsize\textcolor{gray}{[CVPR2025]}}  & \secondcolor{33.28} & \secondcolor{0.954} & \secondcolor{33.31} & \secondcolor{0.959} & \secondcolor{28.87} & 0.863 & \secondcolor{28.92} & \secondcolor{0.882} & \secondcolor{34.78} & 0.949 & \secondcolor{32.71} & \firstcolor{0.959} & \secondcolor{29.90} & \secondcolor{0.879} \\
Fr-Diff \cite{liu2025frequency} {\scriptsize\textcolor{gray}{[arXiv2025]}} & {32.75}        & \multicolumn{1}{c}{{0.941}}          & {32.54} & \multicolumn{1}{c}{{0.936}} & {28.32}  & \multicolumn{1}{c}{0.849} & {28.37}         & \multicolumn{1}{c}{{0.858}}         & {34.34} & \multicolumn{1}{c}{0.942} & {31.68} & \multicolumn{1}{c}{{0.948}} & {28.91}           & {0.869}         \\
TP-Diff \cite{liu2025learning} {\scriptsize\textcolor{gray}{[ICCV2025]}}                          & {32.56}        & \multicolumn{1}{c}{{0.937}}          & {32.44} & \multicolumn{1}{c}{{0.929}} & {28.17}  & \multicolumn{1}{c}{0.845} & {28.24}         & \multicolumn{1}{c}{{0.851}}         & {34.27} & \multicolumn{1}{c}{0.938} & {31.53} & \multicolumn{1}{c}{{0.945}} & {28.86}           & {0.867}         \\
DehazeSB \cite{lan2025schrodinger} {\scriptsize\textcolor{gray}{[ICCV2025]}} & {32.17}        & \multicolumn{1}{c}{{0.934}}          & {32.03} & \multicolumn{1}{c}{{0.925}} & {28.02}  & \multicolumn{1}{c}{0.840} & {28.19}         & \multicolumn{1}{c}{{0.848}}         & {34.25} & \multicolumn{1}{c}{0.930} & {31.55} & \multicolumn{1}{c}{{0.946}} & {28.78}           & {0.861}         \\

\textbf{RGSUD (Ours)} & \firstcolor{34.41} & \firstcolor{0.967} & \firstcolor{33.89} & \firstcolor{0.961} & \firstcolor{29.07} & \firstcolor{0.866} & \firstcolor{29.59} & \firstcolor{0.898} & \firstcolor{35.50} & \firstcolor{0.957} & \firstcolor{32.88} & \secondcolor{0.955} & \firstcolor{30.54} & \firstcolor{0.897} \\ 
\bottomrule
\end{tabular}
\vspace{-0.2in}}
\label{Quantitation-1}
\end{table*}

\begin{figure*}[!ht]
	\centering
	\includegraphics[width=1.0\textwidth]{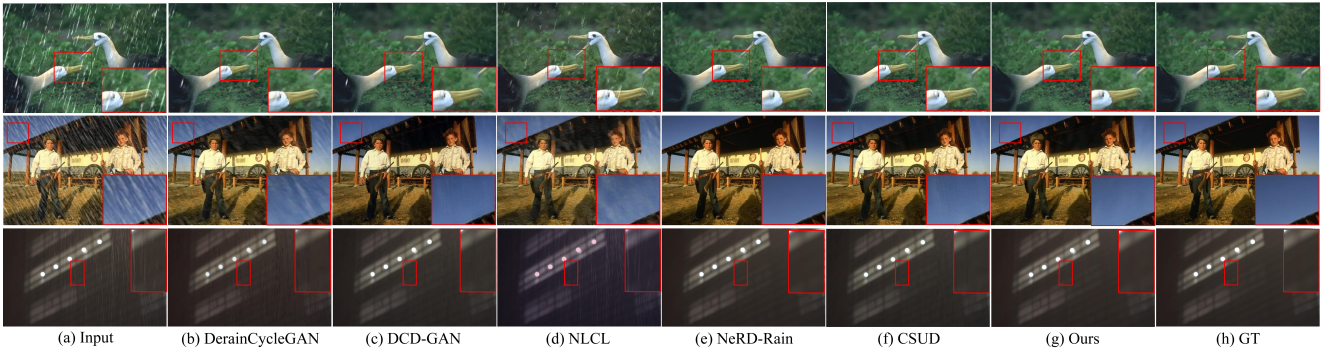}
	\vspace{-6mm}
	\caption{Qualitative deraining performance comparisons on Rain200L, DID-Data, and RealRain1K-L datasets. Our RGSUD achieves competitive visual results comparable to the SOTA supervised method NeRD-Rain-S.}
	\label{fig3}
	\vspace{-3mm}
\end{figure*}

\subsection{Optimization Objective}

\noindent {\textbf{\ding{113} Optimization Objective of Stage One.}}
We simultaneously train a discriminator ${D}$ to distinguish whether a given rainy image $O_{r}^1 $ is synthesized by ${G}$ or sampled from the real-world dataset $O_r$, and the adversarial loss $\mathcal{L}_{adv1}$ is constrained between them. Furthermore, motivated by existing methods \cite{zhu2017unpaired,lin2024re}, the other three adversarial losses can be constructed similarly by constraining the current generated image and $O_r$:
\begin{equation}
\begin{split}
\mathcal{L}_{adv} = \mathcal{L}_{adv1}+\mathcal{L}_{adv2}+\mathcal{L}_{adv3}+\mathcal{L}_{adv4}.
\end{split}
\label{eq1}
\end{equation}

In addition, due to the high difficulty and instability of training an unsupervised framework, we incorporate the PSNR loss and SSIM loss for the derainer:
\begin{equation}
\begin{split}
\mathcal{L}_{Derainer} = \mathcal{L}_{PSNR}(B_u,B_u^{1})+\mathcal{L}_{SSIM}(B_u,B_u^{1}),
\end{split}
\label{eq6}
\end{equation}
the full objective function for stage one $\mathcal{L}_{s1}$ is a weighted sum of the above losses, formulated as:
\begin{equation}
\begin{split}
\mathcal{L}_{s1}=\mathop{min}_{{G}}\mathop{max}_{{D}}\mathcal{L}_{adv}+\lambda_1\mathcal{L}_{Derainer},
\end{split}
\label{eq7}
\end{equation}
where $\lambda_1$ denotes the hyperparameters for $\mathcal{L}_{Der}$.

\noindent {\textbf{\ding{113} Optimization Objective of Stage Two.}}
Supervise learning can  be formulated as a MAP problem:
\begin{equation}
\begin{split}
\mathop{max}_{\theta}\ p(\theta|{O},{B})\propto\ \mathop{max}_{\theta}\ p({B}|{O},\theta)\cdot p(\theta),
\end{split}
\label{eq3}
\end{equation}
Eqn. (\ref{eq3}) can be reformulated as the minimization problem expressed in Eqn. (\ref{eq4}):
\begin{equation}
\begin{split}
\mathop{argmin}_{\theta}\ \overbrace{||{B}-\mathcal{F}_{\theta}({O})||^2_F}^{Data\ Consistency}+\lambda \overbrace{P(\theta)}^{Regularization},
\end{split}
\label{eq4}
\end{equation}
where ${O}$ and ${B}$ represent paired rainy and clean images respectively, while ${\theta}$ denotes the parameters of the target network. Inspired by NLCL \cite{ye2022unsupervised}, the adversarial loss can be regarded as a regularization term. The unsupervised deraining task inherently lacks a data consistency term. We propose to compensate for the absence of the data consistency term by leveraging our previously recycled rewards, as

\begin{equation}
\begin{split}
\mathcal{L}_{re}=||{B}_{rw}-{B}_{r}||^2_F,
\end{split}
\label{eq5}
\end{equation}
where ${B}_{rw}$ denotes reward. 
Compared to prior unsupervised methods solely reliant on regularization, our approach offers enhanced fidelity and clearer optimization trajectories, ensuring a precise distribution match between the restored outputs and clear counterparts.
Finally, we obtain the total loss to train our framework in the two stage: 
\begin{equation}
\begin{split}
\mathcal{L}_{total}=\mathcal{L}_{s1}+\lambda_2 \mathcal{L}_{re},
\end{split}
\label{total}
\end{equation}
where $\lambda_2$ denotes the hyperparameters for $\mathcal{L}_{re}$.

\section{Experiments}

\begin{figure*}[t]
	\centering
	\includegraphics[width=1.0\textwidth]{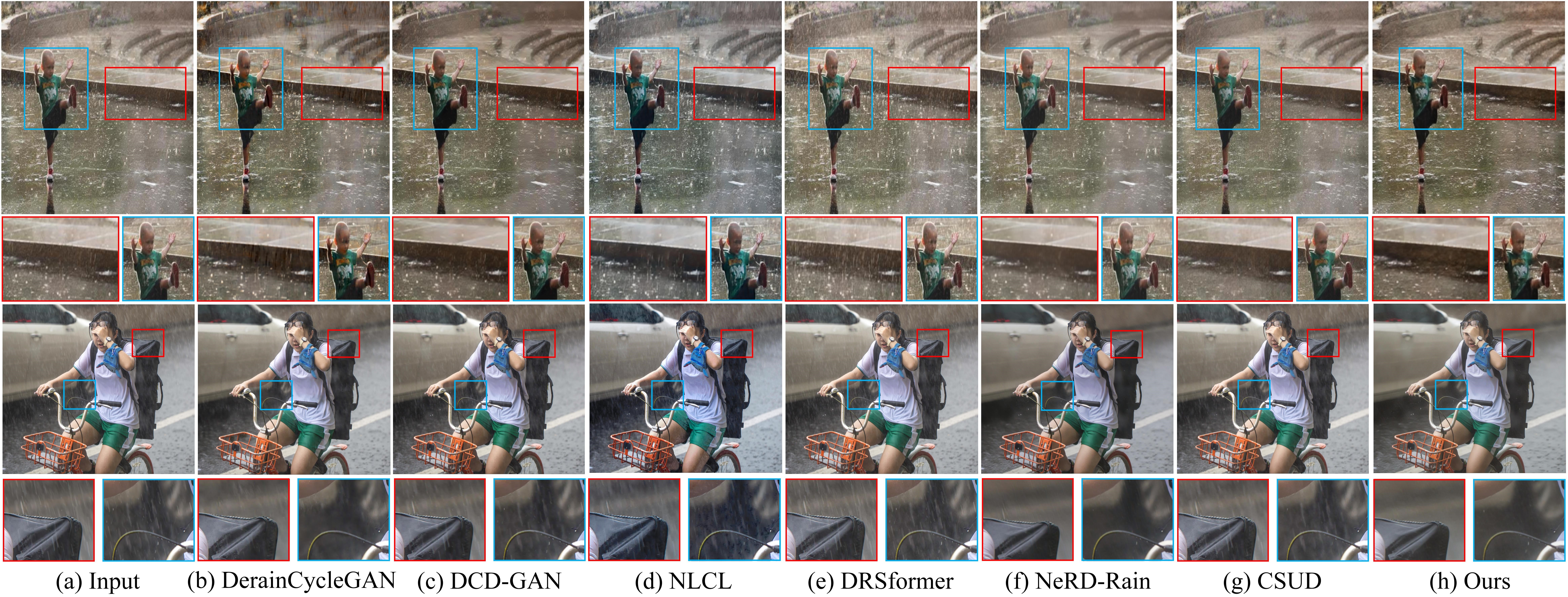}
	\vspace{-6mm}
	\caption{Compared with derained results on the SIRR datasets, and Real3000 datasets, RGSUD recovers clearer images. In outdoor scenarios, our method preserves significant facial details while effectively removing rain.}
	\label{fig4}
	\vspace{-5mm}
\end{figure*}

\subsection{Datasets and Metrics} 

\noindent {\textbf{\ding{113} Datasets.}}
(1) \textcolor{gray}{{\textit{\textbf{Paired Datasets}}}}: Four well-known synthetic paired datasets (\textit{i.e.,} Rain200L \cite{yang2017deep}, Rain100L \cite{yang2017deep}, DID-Data \cite{zhang2018density}, and DDN-Data \cite{liu2020deep}), two real-world paired datasets (\textit{i.e.,} SPA-Data \cite{wang2019spatial} and RealRain1K-L \cite{li2022toward}), and a nighttime paired dataset Night-Rain~\cite{dong2025channel}. (2) \textcolor{gray}{{\textit{\textbf{Unpaired Datasets}}}}: including two real scenario datasets without ground-truth (\textit{i.e.,} SIRR \cite{wei2019semi} and Real3000 \cite{liu2021unpaired}).

\noindent {\textbf{\ding{113} Metrics.}}
(1) \textcolor{gray}{\textit{\textbf{Object Reference IQA Metrics}}}: PSNR and SSIM %
(2) \textcolor{gray}{\textit{\textbf{No-reference IQA Metrics}}}: CLIP-IQA \cite{wang2023exploring}, NIMA \cite{talebi2018nima}, MUSIQ \cite{ke2021musiq}, Q-Align \cite{wu2023q1}, DeQA-Score \cite{you2025teaching}. 

 
\subsection{Implementation Details} Our methodology is executed using the PyTorch platform, conducted on a system powered by four NVIDIA Tesla V100 GPUs. For the training phase, we use the Adam optimizer with $\beta_1=0.9$, $\beta_2=0.999$, and an initial learning rate of $2\times 10^{-4}$. All training images are randomly cropped into 256 × 256 patches in \textbf{an unpaired manner}. 
The hyperparameters $\lambda_1$, $\lambda_2$ are set to 1.0 and 0.8, respectively.

\subsection{Comparison Results on Paired Datasets}
\noindent {\textbf{\ding{113} Quantitative Assessment.}}
We compare \textbf{RGSUD} with unsupervised deraining methods from the most recent years, including CycleGAN \cite{zhu2017unpaired}, DerainCycleGAN \cite{wei2021deraincyclegan}, NLCL \cite{ye2022unsupervised}, DCD-GAN \cite{chen2022unpaired}, and CSUD \cite{dong2025channel}. %
Given the paucity of unsupervised deraining models, we extend our evaluation to include competitive unpaired restoration paradigms (Fr-Diff~\cite{liu2025frequency}, Tp-Diff \cite{liu2025learning}, and DehazeSB \cite{lan2025schrodinger}) by re-train them for the deraining task to establish a robust baseline. %
We also list some SOTA supervised methods. Table \ref{Quantitation-1} presents the performance metrics on synthetic and real datasets. It can be observed that our Method RGSUD significantly outperforms other unsupervised approaches on most datasets, while remaining competitive on the nighttime dataset. Specifically, compared to CSUD, our RGSUD achieves $1.13$ dB and $0.72$ dB improvement in PSNR on Rain100L and SPA-Data datasets, respectively. What's more, our RGSUD even achieves comparable performance to several supervised methods~\cite{11371584}.

\begin{table}[!th]
            \vspace{-0mm}
	\centering
    \caption{Quantitative comparison of deraining methods on both \textcolor{gray}{{\textit{\textbf{Synthetic Paired Datasets}}}} and \textcolor{gray}{{\textit{\textbf{Real-world Paired Datasets}}}} using \textcolor{gray}{\textit{\textbf{Subject Perceptual Metrics}}}.}
    	\vspace{-2.5mm}
     \renewcommand{\arraystretch}{1.1}
\resizebox{1.0\linewidth}{!}{
\begin{tabular}{llccccc}
\toprule
\multirow{2}{*}{Datasets} & \multirow{2}{*}{Metrics} & \multicolumn{1}{c}{\textcolor{gray}{{\textit{\textbf{Supervised}}}}} & \multicolumn{4}{c}{\textcolor{gray}{{\textit{\textbf{Unsupervised}}}}} \\ 
\cmidrule(lr){3-3}
\cmidrule(lr){4-7}
~ & ~ & NeRD-Rain \cite{chen2024bidirectional} & NLCL \cite{ye2022unsupervised} & DCD-GAN \cite{chen2022unpaired} & CSUD \cite{dong2025channel}& Ours \\ 
\midrule
\multicolumn{1}{l}{\multirow{4}{*}{Rain100L \cite{yang2017deep}}}     & CLIP-IQA~\cite{wang2004image}$\uparrow$     & 0.592    & 0.273   &   0.372    & \secondcolor{0.445}   & \firstcolor{0.494}   \\
\multicolumn{1}{c}{}                              & MUSIQ~\cite{ke2021musiq}$\uparrow$ & 66.22  & 50.35 & 54.86     & \secondcolor{57.93} & \firstcolor{62.98} \\
\multicolumn{1}{c}{}                              & Q-Align~\cite{wu2023q1}$\uparrow$      &  4.871   & 3.243   & 3.715      & \secondcolor{3.987}   & \firstcolor{4.358}   \\
\multicolumn{1}{c}{}                              & DeQA-Score \cite{you2025teaching}$\uparrow$   &  4.452   & 2.973   & 3.345      & \secondcolor{3.541}   & \firstcolor{3.873}   \\ 
\midrule
\multicolumn{1}{l}{\multirow{4}{*}{DDN-Data \cite{zhang2018density}}}     & CLIP-IQA~\cite{wang2004image}$\uparrow$     &  0.512   & 0.252   & 0.281      & \secondcolor{0.325}   & \firstcolor{0.357}   \\
\multicolumn{1}{c}{}                              & MUSIQ~\cite{ke2021musiq}$\uparrow$        &  65.19 & 44.39 & 47.68    & \secondcolor{49.14} & \firstcolor{55.14} \\
\multicolumn{1}{c}{}                              & Q-Align~\cite{wu2023q1}$\uparrow$      &  4.492   & 2.315   & 3.104      & \secondcolor{3.235}   & \firstcolor{3.496}   \\
\multicolumn{1}{c}{}                              & DeQA-Score \cite{you2025teaching}$\uparrow$   &  4.363   & 2.857   & 3.132      & \secondcolor{3.423}   & \firstcolor{3.655}   \\ 
\midrule
\multicolumn{1}{l}{\multirow{4}{*}{SPA-Data \cite{wang2019spatial}}}     & CLIP-IQA~\cite{wang2004image}$\uparrow$     &  0.62   & 0.351    & 0.422      & \secondcolor{0.454}   & \firstcolor{0.51}   \\
\multicolumn{1}{c}{}                              & MUSIQ~\cite{ke2021musiq}$\uparrow$        &  69.32 & 52.79 & 55.33    & \secondcolor{57.44} & \firstcolor{59.93} \\
\multicolumn{1}{c}{}                              & Q-Align~\cite{wu2023q1}$\uparrow$      &  4.733   & 3.234   & 3.437      & \secondcolor{3.822}   & \firstcolor{4.121}   \\
\multicolumn{1}{c}{}                              & DeQA-Score \cite{you2025teaching}$\uparrow$   &  4.582   & 3.236   & 3.468      & \secondcolor{3.743}   & \firstcolor{3.981}   \\ 
\midrule
\multicolumn{1}{l}{\multirow{4}{*}{RealRain1K-L \cite{li2022toward}}} & CLIP-IQA~\cite{wang2004image}$\uparrow$     &   0.544   & 0.191   & 0.334      & \secondcolor{0.372}   & \firstcolor{0.413}   \\
\multicolumn{1}{c}{}                              & MUSIQ~\cite{ke2021musiq}$\uparrow$        &  63.59 & 48.34 & 51.76    & \secondcolor{53.63} & \firstcolor{55.73} \\
\multicolumn{1}{c}{}                              & Q-Align~\cite{wu2023q1}$\uparrow$      &  4.753   & 3.153   & 3.346      & \secondcolor{3.588}   & \firstcolor{3.957}   \\
\multicolumn{1}{c}{}                              & DeQA-Score \cite{you2025teaching}$\uparrow$   &  4.792   & 3.126   & 3.497      & \secondcolor{3.652}   & \firstcolor{3.755}   \\ 
\bottomrule
\end{tabular}
	}

	\label{perceptual}	
    \vspace{-4mm}
\end{table}    

\begin{table}[!th]

	\centering
    \caption{Quantitative comparison of deraining methods on \textcolor{gray}{{\textit{\textbf{Real-world Unpaired Datasets}}}} using \textcolor{gray}{\textit{\textbf{No-reference IQA Metrics}}}.}
	\vspace{-2.5mm}
     \renewcommand{\arraystretch}{1.1}
    \resizebox{1.0\linewidth}{!}{
    \begin{tabular}{llccccccccc}
    \toprule
    \multirow{2}{*}{Datasets} & \multirow{2}{*}{Metrics} & \multicolumn{1}{c}{\textcolor{gray}{{\textit{\textbf{Supervised}}}}} & \multicolumn{4}{c}{\textcolor{gray}{{\textit{\textbf{Unsupervised}}}}} \\ 
    \cmidrule(lr){3-3}
    \cmidrule(lr){4-7}
    ~ & ~ & NeRD-Rain \cite{chen2024bidirectional} & NLCL \cite{ye2022unsupervised} & DCD-GAN \cite{chen2022unpaired} & CSUD \cite{dong2025channel}& Ours \\
    \midrule
    \multicolumn{1}{c}{}         & MUSIQ~\cite{ke2021musiq}$\uparrow$           & 57.30 & 57.55  & 57.09   & \secondcolor{58.50}   & \firstcolor{59.08} \\
    \multicolumn{1}{l}{SIRR \cite{wei2019semi}}     & DACLIP-IQA~\cite{luo2024controlling}$\downarrow$    &  0.172 & 0.144   & \secondcolor{0.052}    & 0.072    & \firstcolor{0.036}  \\
    \multicolumn{1}{c}{}         & NIMA \cite{talebi2018nima}$\uparrow$            & 0.102  & 0.103   & 0.108    & \secondcolor{0.109}    & \firstcolor{0.112}  \\ 
    \midrule
    \multicolumn{1}{c}{}         & MUSIQ~\cite{ke2021musiq}$\uparrow$           & 60.52 & 59.66  & 59.95   & \secondcolor{60.92}   & \firstcolor{61.64} \\
    \multicolumn{1}{l}{Real3000 \cite{liu2021unpaired}} & DACLIP-IQA~\cite{luo2024controlling}$\downarrow$    & 0.072  & 0.067   & \secondcolor{0.041}    & 0.042    & \firstcolor{0.018}  \\
    \multicolumn{1}{c}{}         & NIMA~\cite{talebi2018nima}$\uparrow$            &  0.110 & 0.102   & 0.120    & \secondcolor{0.170}    & \firstcolor{0.193}  \\ 
    \bottomrule
    \end{tabular}
	}
	\label{real-world rainy}	
    \vspace{-6mm}
\end{table}    

\noindent {\textbf{\ding{113} Qualitative Comparison.}}
 As shown in Figure \ref{fig3}, our RGSUD achieves better deraining results than other unsupervised methods. In visualizations of nighttime rain removal, our method achieves fewer residual rain streaks. This is attributed to the enhanced perceptual performance of our method, guided by the VLM-IQA, which leverages human visual perception. As shown in Table \ref{perceptual}, our method significantly outperforms other unsupervised methods.

\subsection{Comparison Results on Unpaired Datasets}
\noindent {\textbf{\ding{113} Quantitative Assessment.}}
To evaluate the effectiveness of RGSUD in real-world scenarios, we conduct comprehensive comparisons with SOTA supervised and unsupervised methods on the SIRR dataset and the Real3000 dataset. 
All methods are trained on the Rain100L dataset. The unpaired unsupervised method utilizes only the clean images from Rain100L. Supervised methods demonstrate superior performance on benchmark datasets. But as shown in Table \ref{real-world rainy}, our RGSUD almost comprehensively surpasses existing unsupervised deraining methods. This strongly demonstrates the excellent generalization ability of our method.

\noindent {\textbf{\ding{113} Qualitative Comparison.}}
As shown in Figure \ref{fig4}, the performance of some advanced supervised methods like NeRD-Rain-S and DRSformer is poor when directly applied to real-world rainy data. In contrast, our RGSUD achieves better visual results in the real world. Our method removes the maximum amount of rain streaks while preserving the texture details of the background.

\subsection{Generalization on Downstream Tasks}
\begin{figure}[!t]
\vspace{-0mm}
    \centering
    \includegraphics[width=0.47\textwidth]{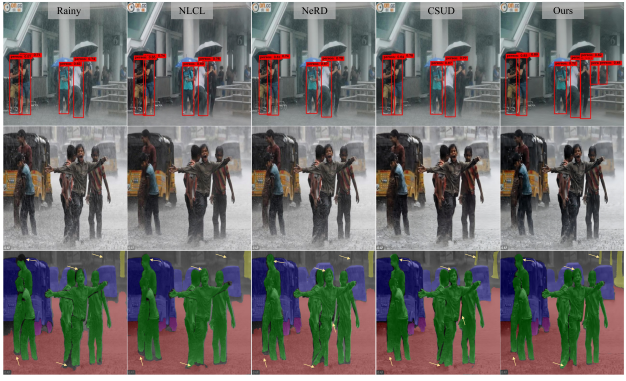}
    \caption{The application performance of derained images in downstream tasks.} 
    \label{downstream} 
    \vspace{-6mm}
\end{figure}

This section validates the adaptability of deraining methods to downstream tasks. For a fair comparison, we directly process the derained results using the pre-trained weights of DeepLabv3+~\cite{chen2017rethinking} and YOLOv8 \cite{ultralytics_yolov8_2023}. Figure \ref{downstream} presents the visual comparisons of semantic segmentation and object detection, respectively. The results of our method detect and recognize more targets, and in semantic segmentation, the generated maps clearly distinguish between regions of different categories.

\subsection{Ablation Analysis}

\subsubsection{Effectiveness of SR Strategy} 
\noindent {\textbf{\ding{113} SR Strategy.}}
To verify the effectiveness of the SR strategy, we conduct ablation experiments on the benchmark datasets. As shown in Table \ref{table10}, compared to the baseline, the effect on PSNR is improved by up to $1$ dB on Rain100L, DID-Data, and RealRain1K-L datasets with the SR strategy. The above ablation results strongly verify the superiority and importance of the SR strategy.

\noindent {\textbf{\ding{113} Derainer Transferability.}}
The Derainer in our backbone network is replaceable, and this section verifies the scalability of our method.
We replace different supervised deraining network architectures as the derainer, including Restormer \cite{zamir2022restormer}, DRSformer~\cite{chen2023learning}, and NeRD-Rain~\cite{chen2024bidirectional}. The framework still adopts the RGSUD with different deraining networks. 
Table \ref{table3} illustrates that our proposed SR strategy is effective for these networks. For instance, after applying the SR strategy to Restormer, the PSNR improvements on Rain200L are $0.42$ dB. For DRSformer and NeRD-Rain, the gains are $0.69$ dB and $0.78$ dB, respectively. Consequently, the results demonstrate that our framework and SR strategy have strong transferability~\cite{11171582}.

\begin{table}[!t]
    \renewcommand{\arraystretch}{1.2}
	\centering
    \tiny
	\caption{The effectiveness of the SR strategy of our method.}
	\vspace{-2.5mm}
	\resizebox{1.0\linewidth}{!}{
    \begin{tabular}{lcccc}
    \toprule
    \multicolumn{1}{l}{\multirow{2}{*}{Datasets}} & \multicolumn{2}{c}{\textcolor{gray}{{\textit{\textbf{without SR Strategy}}}}} & \multicolumn{2}{c}{\textcolor{gray}{{\textit{\textbf{with SR Strategy}}}}} \\ 
    \cmidrule(lr){2-3}
    \cmidrule(lr){4-5}
    ~ & PSNR$\uparrow$ & SSIM$\uparrow$ & PSNR$\uparrow$ & SSIM$\uparrow$ \\ 
    \midrule
    Rain100L \cite{yang2017deep} & 33.04 & 0.949 & 34.41 {\tiny\textcolor{gray}{(+1.37)}} & 0.967 {\tiny\textcolor{gray}{(+0.018)}} \\
    Rain200L \cite{yang2017deep} & {32.93} & {0.948} & {33.89} {\tiny\textcolor{gray}{(+0.96)}} & {0.961} {\tiny\textcolor{gray}{(+0.013)}} \\
    DID-Data \cite{zhang2018density} & 28.02                                            & 0.845 & 29.07 {\tiny\textcolor{gray}{(+1.05)}} & 0.866 {\tiny\textcolor{gray}{(+0.021)}} \\
    DDN-Data \cite{liu2020deep} & {28.76} & {0.875} & {29.59} {\tiny\textcolor{gray}{(+0.83)}} & {0.898} {\tiny\textcolor{gray}{(+0.023)}} \\
    SPA-Data \cite{wang2019spatial} & 34.80 & 0.946 & 35.50 {\tiny\textcolor{gray}{(+0.70)}} & 0.957 {\tiny\textcolor{gray}{(+0.011)}} \\
    Night-Rain \cite{dong2025channel} & {29.88} & {0.876} & {30.54} {\tiny\textcolor{gray}{(+0.66)}} & {0.897} {\tiny\textcolor{gray}{(+0.021)}} \\
    RealRain1K-L \cite{li2022toward} & 31.31 & 0.943 & 32.88 {\tiny\textcolor{gray}{(+1.57)}} & 0.955 {\tiny\textcolor{gray}{(+0.012)}} \\ 
    \bottomrule
    \end{tabular}

	}
        \vspace{-4mm}
	\label{table10}	
\end{table} 

\begin{table}[!t]
        \renewcommand{\arraystretch}{1.1}
	\centering
	\caption{The effects of SR strategy for different derainers.}
	\vspace{-2.5mm}
	\resizebox{1.0\linewidth}{!}{
            \begin{tabular}{lccccccc}
            \toprule
            \multirow{2}{*}{Derainers}  & \multirow{2}{*}{SR Strategy} & \multicolumn{2}{c}{Rain200L \cite{yang2017deep}} & \multicolumn{2}{c}{SPA-Data \cite{wang2019spatial}} & \multicolumn{2}{c}{RealRain1K-L \cite{li2022toward}} \\ \cline{3-8} 
                                       &                                & PSNR$\uparrow$          & SSIM$\uparrow$          & PSNR$\uparrow$        & SSIM$\uparrow$        & PSNR$\uparrow$             & SSIM$\uparrow$            \\ 
            \midrule
            \multirow{2}{*}{Restormer \cite{zamir2022restormer}} & No                             &  32.61             & 0.936             & 34.29           & 0.937           &  31.19               & 0.938               \\
                                       & {Yes}\cellcolor[HTML]{E6E6E6}                             &  {\textbf{33.03}}\cellcolor[HTML]{E6E6E6}              & {\textbf{0.940}}\cellcolor[HTML]{E6E6E6}              & {\textbf{35.16}}\cellcolor[HTML]{E6E6E6}            & {\textbf{0.945}}\cellcolor[HTML]{E6E6E6}            &  {\textbf{32.28}}\cellcolor[HTML]{E6E6E6}                & {\textbf{0.943}}\cellcolor[HTML]{E6E6E6}                \\ 
            \midrule
            \multirow{2}{*}{DRSformer \cite{chen2023learning}} & No                             &  32.79             & 0.938             & 34.52           & 0.931           &   31.09              &  0.934              \\
                                       & {Yes}\cellcolor[HTML]{E6E6E6}                             &  {\textbf{33.48}}\cellcolor[HTML]{E6E6E6}              & {\textbf{0.951}}\cellcolor[HTML]{E6E6E6}              & {\textbf{35.28}}\cellcolor[HTML]{E6E6E6}            & {\textbf{0.949}}\cellcolor[HTML]{E6E6E6}            &   {\textbf{32.36}}\cellcolor[HTML]{E6E6E6}               &  {\textbf{0.948}}\cellcolor[HTML]{E6E6E6}               \\ 
            \midrule
            \multirow{2}{*}{NeRD-Rain \cite{chen2024bidirectional}} & No                             &  32.85             & 0.942             & 34.74           & 0.948           &  31.17               & 0.942               \\
                                       & {Yes}\cellcolor[HTML]{E6E6E6}                             &  {\textbf{33.63}}\cellcolor[HTML]{E6E6E6}              & {\textbf{0.954}}\cellcolor[HTML]{E6E6E6}              & {\textbf{35.43}}\cellcolor[HTML]{E6E6E6}            & {\textbf{0.954}}\cellcolor[HTML]{E6E6E6}            &  {\textbf{32.47}}\cellcolor[HTML]{E6E6E6}                & {\textbf{0.951}}\cellcolor[HTML]{E6E6E6}                \\ 
            \midrule
            \multirow{3}{*}{NAFNet \cite{chen2022simple}}    & No                             & 32.93              & 0.948             & 34.80           & 0.946           &  31.31               &  0.943              \\
                                       & {Yes}\cellcolor[HTML]{E6E6E6}                             & {\textbf{33.89}}\cellcolor[HTML]{E6E6E6}               & {\textbf{0.959}}\cellcolor[HTML]{E6E6E6}              & {\textbf{35.50}}\cellcolor[HTML]{E6E6E6}            & {\textbf{0.957}}\cellcolor[HTML]{E6E6E6}            &  {\textbf{32.88}}\cellcolor[HTML]{E6E6E6}                &  {\textbf{0.955}}\cellcolor[HTML]{E6E6E6}               \\ 
            \bottomrule
            \end{tabular}

	}
	\label{table3}	
\vspace{-4mm}
\end{table}    

\begin{table}[!t]
        \renewcommand{\arraystretch}{1.1}
	\centering
	\caption{The effectiveness of the SR strategy for different unsupervised deraining methods.}
	\vspace{-2.5mm}
	\resizebox{1.0\linewidth}{!}{
            \begin{tabular}{lccccccc}
            \toprule
            \multirow{2}{*}{Methods}         & \multirow{2}{*}{SR Strategy} & \multicolumn{2}{c}{Rain100L \cite{yang2017deep}} & \multicolumn{2}{c}{DID-Data \cite{zhang2018density}} & \multicolumn{2}{c}{RealRain1K-L \cite{li2022toward}} \\ \cline{3-8} 
                                             &                              & PSNR$\uparrow$          & SSIM $\uparrow$        & PSNR $\uparrow$         & SSIM$\uparrow$         & PSNR$\uparrow$            & SSIM$\uparrow$           \\ 
            \midrule
            \multirow{2}{*}{\begin{tabular}[c]{@{}l@{}}Derain-\\ CycleGAN \cite{wei2021deraincyclegan}\end{tabular}} & No                           & 32.31         & 0.946        & 28.43         & 0.864        & 28.16           & 0.901          \\
                                             & {Yes}\cellcolor[HTML]{E6E6E6}                           & {\textbf{32.93}}\cellcolor[HTML]{E6E6E6}          & {\textbf{0.949}}\cellcolor[HTML]{E6E6E6}         & {\textbf{28.57}}\cellcolor[HTML]{E6E6E6}          & {\textbf{0.871}}\cellcolor[HTML]{E6E6E6}         & {\textbf{28.31}}\cellcolor[HTML]{E6E6E6}            & {\textbf{0.905}}\cellcolor[HTML]{E6E6E6}           \\ 
                                             \midrule
            \multirow{2}{*}{NLCL \cite{ye2022unsupervised}}            & No                           & 27.86         & 0.852        & 25.89         & 0.813        & 23.06           & 0.832          \\
                                             & {Yes}\cellcolor[HTML]{E6E6E6}                           & {\textbf{27.92}}\cellcolor[HTML]{E6E6E6}          & {\textbf{0.856}}\cellcolor[HTML]{E6E6E6}         & {\textbf{25.94}}\cellcolor[HTML]{E6E6E6}          & {\textbf{0.817}}\cellcolor[HTML]{E6E6E6}         & {\textbf{23.13}}\cellcolor[HTML]{E6E6E6}            & {\textbf{0.838}}\cellcolor[HTML]{E6E6E6}           \\ 
                                             \midrule
            \multirow{2}{*}{DCD-GAN \cite{chen2022unpaired}}         & No                           & 31.82         & 0.941        & 28.64         & 0.862        & 30.49           & 0.939          \\
                                             & {Yes}\cellcolor[HTML]{E6E6E6}                           & {\textbf{32.23}}\cellcolor[HTML]{E6E6E6}          & {\textbf{0.945}}\cellcolor[HTML]{E6E6E6}         & {\textbf{28.73}}\cellcolor[HTML]{E6E6E6}          & {\textbf{0.864}}\cellcolor[HTML]{E6E6E6}         & {\textbf{30.77}}\cellcolor[HTML]{E6E6E6}            & {\textbf{0.945}}\cellcolor[HTML]{E6E6E6}           \\ 
                                             \midrule
            \multirow{2}{*}{CSUD \cite{dong2025channel}}            & No                           & 33.28         & 0.954        & 28.87         & 0.863        & 32.71           & 0.959          \\
                                             & {Yes}\cellcolor[HTML]{E6E6E6}                          & {\textbf{33.92}}\cellcolor[HTML]{E6E6E6}          & {\textbf{0.957}}\cellcolor[HTML]{E6E6E6}         & {\textbf{28.95}}\cellcolor[HTML]{E6E6E6}          & {\textbf{0.865}}\cellcolor[HTML]{E6E6E6}         & {\textbf{32.87}}\cellcolor[HTML]{E6E6E6}            & {\textbf{0.963}}\cellcolor[HTML]{E6E6E6}           \\ 
                                             \bottomrule
            \end{tabular}

	}
    \vspace{-6mm}
	\label{table4}	
\end{table}    

\noindent {\textbf{\ding{113} Plug-in Ability.}}
Furthermore, we validate the applicability of the proposed training strategy to other unsupervised deraining methods. Specifically, we utilize recycling to recover the rewards during the training process and continue training with a self-reinforcement constraint for an additional 3 hours afterward. As shown in Table \ref{table4}, after applying the SR strategy to DCD-GAN~\cite{chen2022unpaired} and CSUD~\cite{dong2025channel}, the PSNR improvements on Rain100L are $0.41$ dB and $0.68$ dB, respectively. This implies that our SR strategy facilitates a more accurate and compact convergence in the restoration process. However, suboptimal initial deraining results yield inadequate reward, directly compromising subsequent SR strategy performance. Consequently, the SR strategy demonstrates limited performance improvement for NLCL~\cite{ye2022unsupervised}.

\subsubsection{Reliable Metric Selection.} Intuitively, arbitrary non-reference image quality assessment can be incorporated into the dynamic reward recycling mechanism. Therefore, finding the best possible NR-IQA for image deraining is worth considering. We conduct experimental analysis on three metrics: MUSIC~\cite{ke2021musiq}, NIMA~\cite{talebi2018nima}, and CLIP-IQA~\cite{wang2023exploring}. As shown in Table \ref{impact}, the DACLIP-IQA~\cite{luo2024controlling}- based reward recycling mechanism identifies superior rewards, effectively facilitating the implementation of the SR strategy.


\subsubsection{Effect of the Weight Coefficients of Loss Function.} 
We conduct extensive ablation experiments on the weight coefficients of $\mathcal{L}_{Der}$ and $\mathcal{L}_{re}$ in Eqn.(\ref{total}). %
As shown in Table~\ref{table6}, aside from our selected optimal combination, our method achieves enhanced deraining performance with SR strategy integration, demonstrating the effectiveness of our RGSUD framework.

\subsubsection{Model Complexity.} 
In this section, we will analyze the model's complexity and inference speed. DACLIP-IQA~\cite{luo2024controlling} is only used during the training process, and its internal parameters do not participate in gradient backpropagation updates.  As shown in Table \ref{table7}, our method achieves the best performance in terms of ``FLOPs'', `Memory Usage'', and ``Inference Time''. 



\subsubsection{Adversarial Loss and DEM.} 
As shown in Table \ref{DEM}, comparing V1 and V3 indicates that adding $\mathcal{L}_{adv}$ to the framework can lead to a more stable training process. 
Furthermore, comparing V2 with V3, we observe that during the SR strategy implementation stage, rewards help DEM acquire more accurate degradation information. This resulted in the synthesis of higher-quality pseudo-paired data, consequently improving the network's capacity to learn the real-world rain distribution.

\begin{table}[!t]
        \renewcommand{\arraystretch}{1.2}
	\centering
	\caption{Impact of image quality assessment on dynamic reward recycling mechanism.}
	        \vspace{-2.5mm}
	\resizebox{1.0\linewidth}{!}{
            \begin{tabular}{lcccccc}
            \toprule
            \multirow{2}{*}{IQA Methods} & \multicolumn{2}{c}{Rain100L \cite{yang2017deep}} & \multicolumn{2}{c}{DID-Data \cite{zhang2018density}} & \multicolumn{2}{c}{RealRain1K-L \cite{li2022toward}} \\
            \cline{2-3}
            \cline{4-5}
            \cline{6-7}
                                         & PSNR $\uparrow$        & SSIM $\uparrow$        & PSNR $\uparrow$        & SSIM $\uparrow$         & PSNR $\uparrow$           & SSIM  $\uparrow$         \\ \midrule
            MUSIC \cite{ke2021musiq} & 33.56         & 0.959         & 28.34         & 0.850         & 31.65           & 0.945          \\
            NIMA \cite{talebi2018nima} & 33.32         & 0.955         & 28.33         & 0.848         & 31.54           & 0.944          \\
            CLIP-IQA \cite{wang2023exploring} & 33.67         & 0.961         & 28.47         & 0.853         & 31.84           & 0.948          \\
            {DACLIP-IQA \cite{luo2024controlling}}\cellcolor{gray!20}                    & {\textbf{34.41}}\cellcolor{gray!20}          & {\textbf{0.967}}\cellcolor{gray!20}          & {\textbf{29.07}}\cellcolor{gray!20}          & {\textbf{0.866}}\cellcolor{gray!20}          & {\textbf{32.88}}\cellcolor{gray!20}            & {\textbf{0.955}} \cellcolor{gray!20}          \\ \bottomrule
            \end{tabular}

	}
	\label{impact}	
    \vspace{-4mm}
\end{table}    

\begin{table}[!t]
        \renewcommand{\arraystretch}{1.1}
	\centering
    \scriptsize
	\caption{Ablation experiments on the weight coefficients of the loss function.}
	\vspace{-2.5mm}
	\resizebox{1.0\linewidth}{!}{
            \begin{tabular}{cccccccc}
            \hline
            \multirow{2}{*}{$\lambda_1$} & \multirow{2}{*}{$\lambda_2$} & \multicolumn{2}{c}{Rain100L \cite{yang2017deep}} & \multicolumn{2}{c}{DID-Data \cite{zhang2018density}} & \multicolumn{2}{c}{RealRain1K-L \cite{li2022toward}} \\
            \cline{3-4}
            \cline{5-6}
            \cline{7-8}
                               &                    & PSNR$\uparrow$          & SSIM$\uparrow$          & PSNR$\uparrow$          & SSIM$\uparrow$          & PSNR$\uparrow$            & SSIM$\uparrow$           \\ \hline
            1.0                & 0.5                & 33.89         & 0.959         & 28.41         & 0.854         & 31.98           & 0.946          \\
            2                  & 0.5                & 33.63         & 0.952         & 28.24         & 0.850         & 31.75           & 0.943          \\
            1.0                & 1.0                & 34.22         & 0.962         & 28.75         & 0.858         & 32.54           & 0.951          \\
            1.0                & 1.5                & 33.98         & 0.960         & 28.53         & 0.856         & 32.43           & 0.949          \\
            {1.0}\cellcolor{gray!20}                   & {0.8}\cellcolor{gray!20}                   & \textbf{34.41}\cellcolor{gray!20}            & \textbf{0.967}\cellcolor{gray!20}            & \textbf{29.07}\cellcolor{gray!20}            & \textbf{0.866}\cellcolor{gray!20}            & \textbf{32.88}\cellcolor{gray!20}              & \textbf{0.955}\cellcolor{gray!20}             \\ \hline
            \end{tabular}

	}
    
	\label{table6}	
    \vspace{-6mm}
\end{table}    

\begin{table}[!t]
        \renewcommand{\arraystretch}{1.2}
	\centering
	\caption{The size of the test image is $256\times256$ pixels. \# FLOP and \# Params represent FLOPs (in G) and the number of trainable parameters (in M), respectively.}
	\vspace{-2.5mm}
	\resizebox{1.0\linewidth}{!}{
            \begin{tabular}{lccccc}
            
            \hline
            Methods      & DCD-GAN \cite{chen2022unpaired} & DerainCycleGAN \cite{wei2021deraincyclegan} & NLCL \cite{ye2022unsupervised} & CSUD \cite{dong2025channel} & Ours \\ \hline
            \# FLOPs (G)         & 89.2    & 65.8       & 41.6     & \secondcolor{30.4}     & \firstcolor{16.3}     \\
            \# Trainable Params (M)        & 17.4    & \firstcolor{0.2}        & \secondcolor{0.3}      & 31.8     & 29.1     \\ 
            \# Memory Usage (G)   &4.3      &1.3         & 5.7      & \secondcolor{0.56}      &\firstcolor{0.50}   \\
            \# Inference Time (s) &0.73     & 0.64       &  0.87    & \secondcolor{0.15}      &\firstcolor{0.11}\\ \hline
            \end{tabular}

	}
        \vspace{-4.5mm}
	\label{table7}	
\end{table}    

\begin{table}[!t]
        \renewcommand{\arraystretch}{1.1}
	\centering
	\caption{V1 and V3 represent network training constrained by $\mathcal{L}_{adv1}$ and $\mathcal{L}_{adv}$. V2 and V3 denote whether reward is included or excluded for auxiliary degradation extraction with DEM.}
	\vspace{-2.5mm}
	\resizebox{1.0\linewidth}{!}{
            \begin{tabular}{c|cccc|cc}
            \hline
            \multirow{2}{*}{Methods} & \multirow{2}{*}{$\mathcal{L}_{adv1}$} & \multirow{2}{*}{$\mathcal{L}_{adv}$} & \multirow{2}{*}{\begin{tabular}[c]{@{}c@{}}DEM \\ without reward\end{tabular}} & \multirow{2}{*}{\begin{tabular}[c]{@{}c@{}}DEM \\ with reward\end{tabular}} & \multicolumn{2}{c}{Rain100L} \\  
                                     &                    &                    &                    &                    & PSNR $\uparrow$         & SSIM$\uparrow$         \\ \hline
            V1                       & \Checkmark                  & \XSolidBrush                  & \XSolidBrush                  & \Checkmark                  & 29.45         & 0.857        \\
            V2                       & \XSolidBrush                  & \Checkmark                   & \Checkmark                  & \XSolidBrush                 &  32.47        &  0.952       \\ 
            {V3}\cellcolor{gray!20}                        & {\XSolidBrush}\cellcolor{gray!20}                   & {\Checkmark}\cellcolor{gray!20}                   & {\XSolidBrush}\cellcolor{gray!20}                   & {\Checkmark}\cellcolor{gray!20}                   & \textbf{34.41}\cellcolor{gray!20}         & \textbf{0.967}\cellcolor{gray!20}         \\ \hline
            \end{tabular}

	}
	\label{DEM}	
    \vspace{-4mm}
\end{table}    

\subsection{Training stability analysis of DEM.}
\begin{figure}[!t]
    \begin{minipage}{0.21\textwidth}
    \centering
     \captionsetup{font={small}}
    \captionof{table}{Ablation on DEM.}
            \vspace{-3mm}
    \resizebox{1\linewidth}{!}{
 \begin{tabular}{ccccc}
\toprule
\multirow{2}{*}{Dataset} & \multicolumn{2}{c}{w/o DEM} & \multicolumn{2}{c}{Ours} \\ \cline{2-5} 
                         & PSNR$\uparrow$         & SSIM$\uparrow$         & PSNR$\uparrow$        & SSIM$\uparrow$       \\ \hline
Rain100L                 & 30.73        & 0.904        & \firstcolor{33.04}       & \firstcolor{0.949}      \\
DID-Data                 & 27.39        & 0.808        & \firstcolor{28.02}       & \firstcolor{0.845}      \\ \bottomrule
\end{tabular}
    }

    \label{DEM1}
    \end{minipage} 
    \hfill   
    \begin{minipage}{0.26\textwidth}
    \centering

\includegraphics[width=0.9\linewidth]{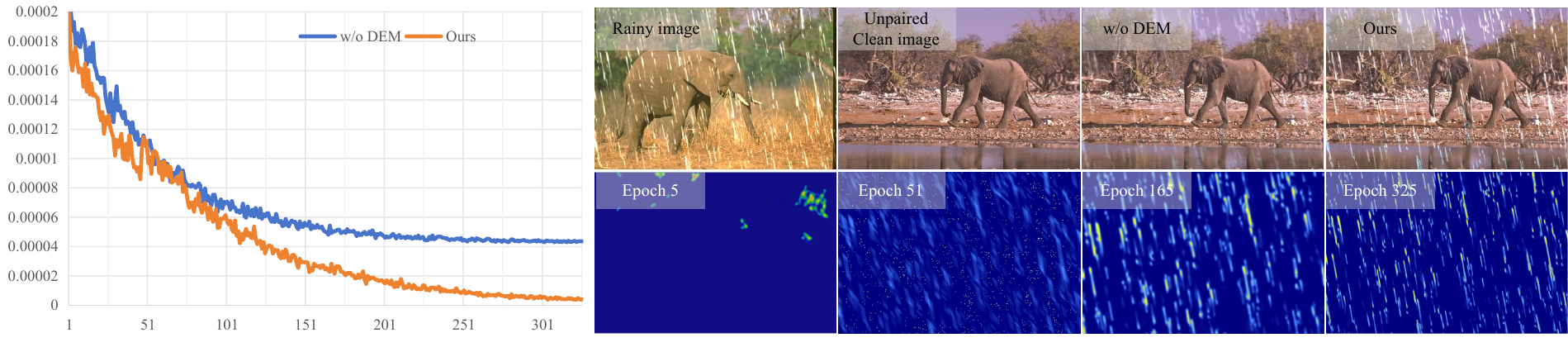}
    \vspace{-3mm}
     \captionsetup{font={small}}
    \caption{Visualization results.}
    \label{DEM2}
    \end{minipage}
    \vspace{-4mm}
\end{figure}

Since the DEM extracts rain-degradation features from real images, it cannot use an explicit loss. We validate the U-net-based DEM by showing that it (1) improves rain removal in Talbe~\ref{DEM1}, (2) accelerates convergence in left of Figure~\ref{DEM2}, (3) yields higher-quality pseudo pairs in in right of Figure~\ref{DEM2}. Stronger backbones may bring further gains, which  we leave for future work.

\begin{table}[!t]
    \centering
    \renewcommand{\arraystretch}{1.2} 
    \caption{he transition condition from Stage I to Stage II.}
    \label{Transitional}
    \vspace{-3mm}
    \resizebox{1.0\linewidth}{!}{
        \begin{tabular}{lcccccccc} 
        \toprule
        \multicolumn{2}{l}{Dataset} & Rain100L & Rain200L & DID-Data & DDN-Data & SPA-Data & RealRain1K-L & Night-Rain \\ \midrule
        \multicolumn{2}{l}{Epochs}  & 357-374  & 363-384  & 267-273  & 251-264  & 4        & 278-304      & 336-367    \\ \midrule
        
        \multirow{2}{*}{Validation} & PSNR$\uparrow$ & 33.04 & 32.93 & 28.02 & 28.76 & 34.80 & 31.31 & 29.88 \\ 
        ~                           & SSIM$\uparrow$ & 0.949 & 0.948 & 0.845 & 0.875 & 0.946 & 0.943 & 0.876 \\ \bottomrule
        \end{tabular}
    }
    \vspace{-6mm}
\end{table} 

\subsection{Transitional conditions.}
As shown in Table~\ref{Transitional}, Stage I→II is triggered based on training epochs and validation PSNR/SSIM: one epoch is a full pass, so larger datasets need fewer epochs, and the transition also reflects rain-streak diversity, occurring once validation PSNR/SSIM stabilizes.


\section{Conclusion}


In this paper, we propose a novel unsupervised deraining framework, \textbf{RGSUD}, that leverages historical optimal deraining results to guide model training. Specifically, we introduce a VLM-based dynamic reward recycling mechanism to select high-quality derained outputs from the training process and use them as rewards. To further enhance optimization, we design a reward-driven loss function, called the self-reinforcement loss, that provides explicit deraining signals, helping the network converge to a compact optimization space. Extensive experiments across multiple synthetic demonstrate that our method significantly improves deraining performance, achieving state-of-the-art results across diverse real-world rainy scenarios. Through extensive ablation studies, we also confirm the effectiveness of our SR strategy, achieving up to 1 dB improvement in PSNR over the baseline on datasets such as Rain100L and RealRain1K-L, further validating the robustness and impact of our approach.
Future work may explore video deraining~\cite{chen2024dual} or night deraining~\cite{lin2024nightrain,jin2023enhancing}, defogging~\cite{jin2022structure} or 3D Reconstruction~\cite{Zhi_2019_CVPR,Zhi_2021_ICCV,9996585,shi2025planerectr++,du2024mose,chen2022single,zhang2023scatterer,10190736}.

{
    \small
    \bibliographystyle{ieeenat_fullname}
    \bibliography{main}
}


\end{document}